\documentclass[letterpaper]{article}
\usepackage{iccc}

\usepackage{svg}
\usepackage{mwe}
\usepackage{times}
\usepackage{helvet}
\usepackage{courier}
\usepackage{mathtools}
\title{StrokeCoder: Path-Based Image Generation from Single Examples using Transformers}
\author{Sabine Wieluch and Friedhelm Schwenker\\
Institute for Neural Information Processing\\
Ulm University\\
89081 Ulm\\
sabine.wieluch@uni-ulm.de, friedhelm.schwenker@uni-ulm.de\\
}
\begin{document} 
\maketitle
\begin{abstract}
\begin{quote}
This paper demonstrates how a Transformer neural network can be used to learn a generative model from a single path-based example image. We further show how a data set can be generated from the example image and how the model can be used to generate a large set of deviated images, which still represent the original image's style and concept.
\end{quote}
\end{abstract}

\section{Introduction}
Hand-drawn sketches are often used to quickly illustrate a scene or easily capture a thought. So a quick drawing often becomes the first step of a large design process and is an essential part in generating new design ideas \cite{suwa1998roles}.\\
To support users in their design processes \cite{do2005design} or to give users the possibility for creative experimentation with sketches, for example with with Casual Creators \cite{compton2019casual}, it will be very interesting to build such supportive tools with the help of generative machine learning.\\
Though for most design process it is not interesting or even not possible to collect a large training data set. For example it is not suitable for a computer game level designer to create thousands of levels to train a generative model, as the data set generation exceeds the generative model's value. For this reason, this work focuses on three main points. First: one-shot learning, where a model is trained by only one or few examples. Second: generative models that produce path-based images, not pixel-based images. For supportive drawing applications and also for other design tasks like in digital fabrication \cite{wang2008retrieving}, it is important to work with path data (for example a laser cutter needs path data to move along a line). Third: generation processes that can derive new work from the training data but preserve the original style.\\
\ \\
Interesting work on sketch image generation has been performed by \cite{singh2017unified}, who also focus on the generation of style preserving derivatives for Co-Creative settings. Also other research groups \cite{yu2020toward,eitz2012humans,yu2016sketch,gasques2019you}, worked on sketch data mainly utilizing Generative Adversarial Networks \cite{radford2015unsupervised,goodfellow2016nips}. Therefore the resulting images are pixel-based and not path-based images.\\
\ \\
Machine learning on path-based data has been performed in related research work. One very well known example is Sketch-RNN \cite{ha2017neural} where a recursive neural net was trained with a large data set of small sketches depicting different objects. Another very interesting approach by \cite{balasubramanian2019teaching} proposes a GAN-like architecture utilizing Long Short-Term Memory (LSTM) neural nets to generate path-based images from large sketch datasets.\\
Recent studies by Xu et al. showed that Transformer neural networks are well suited to be trained on sketch data. Transformers \cite{vaswani2017attention} are state-of-the art architectures for handling sequential data and outperform RNN or LSTM architectures in Natural Language Generation tasks. Xu et al. implemented a graph-based representation for strokes to perform free-hand sketch recognition \cite{xu2019multi,xu2020deep2}.\\
Training and generation of Scalable Vector Graphics has been performed by \cite{lopes2019learned}, where they used several fonts as data sets.\\ 
\ \\
In our last research \cite{wieluch2019dropout}, we examined how Dropout can be used with Generative Adversarial Networks to create different, but coherent images in image-to-image translation tasks. For this research paper we aim to learn a generative model from one single hand-drawn image. We focus on generating diverse images that match the input image's style and concept. The generated images will be especially useful for creative support tools, art or digital fabrication.

\section{Data Structure and Transformer Architecture}
\begin{figure}[h]
	\centering
	\includegraphics[width=4.5cm]{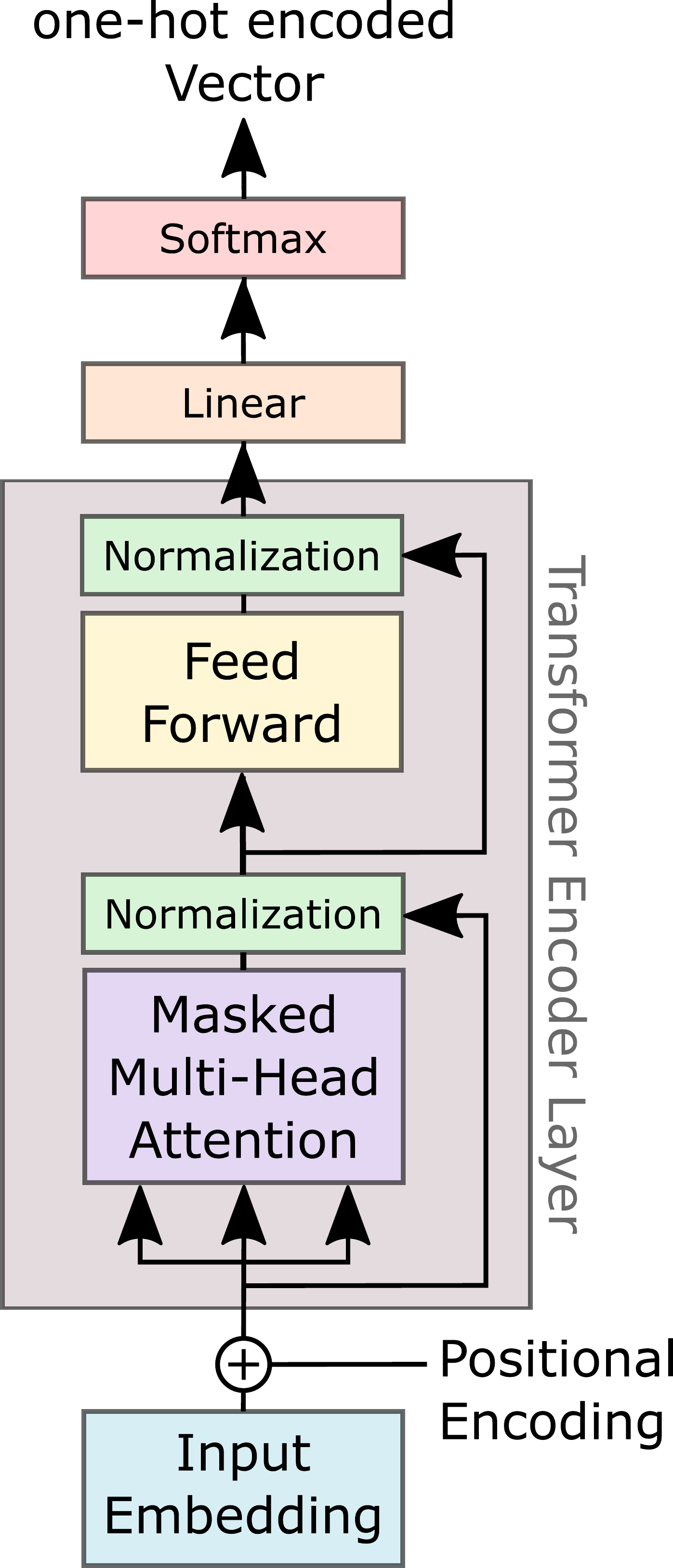}
	\caption{Transformer Encoder let as used in this work: after the input is embedded and receives positional encoding information, it uses multiple self-attention layers and a feed-forward net to process the sequence. Encoder layers can be stacked on top at any number. The Encoder layer output is processed through a linear and a softmax layer to receive the final one-hot encoded vector, which can then be used to read the final move from the embedding.}
	\label{nnet}
\end{figure}

In this research, we aim to learn a neural representation for path-based sketch drawings. A sketch drawing consists of a sequence of pen strokes and each stroke can be approximated by small straight line segments. This sequence of sequences can be easily flattened to one large sequence.\\
Therefore learning a neural representation for sketch drawings can be formulated as a sequence generation task. Transformers \cite{vaswani2017attention} are a new class of neural nets which have been especially useful to the Natural Language Processing community. They are used for sequence-to-sequence translation tasks \cite{devlin2018bert} as well as for text generation \cite{radford2019language}. Also other domains have used Transformers for various tasks like music generation \cite{huang2018music}. \\
A Transformer \cite{vaswani2017attention} consists of an Encoder and a Decoder, as it is usually used for sequence-to-sequence translation tasks. However, in our setting we aim to learn a generative model and therefore we will only use a Transformer Encoder.\\
This seems counterintuitive at first, because for a generation task usually the Decoder would be used. Though Transformers Encoder and Decoder are very similarly constructed and only differ in an additional input from the Encoder to the Decoder in a translation setting. As we only want to generate a sequence, we can discard this input and end up with with the Transformer Encoder.\\
\ \\
The Transformer Encoder consists of multiple layers, which end in a linear and softmax layer. Decoder layers can be stacked on top of each other any number of times.\\
One Decoder layer consists of three sub-layers. The first and second layers are Multi-Head-Attention layers, which are a number of parallel attention layers, whose output is concatenated and finalized with a linear layer. A self-attention layer gives the neural net the ability to focus more on certain moves or ignore other moves in the sequence. A mask is applied to the first attention layer to prevent the neural net from seeing future sequence elements. Each of these two sub-layers end with a layer normalization. The third sub-layer is a feed-forward network, also ending with a layer normalization. Figure \ref{nnet} gives a visual overview of our used architecture.\\
\ \\
The Transformer architecture does not receive information if the positional order of the embedded move sequence. Therefore, additional positional encoding is required. In our research, we use the standard positional encoding defined in \cite{vaswani2017attention}.\\
\ \\
But before we can use the Encoder, the input needs to be prepared accordingly. The sequence of straight lines needs to be converted to a sequence of vectors that can be processed by a neural network. In Natural Language Processing, using word sequences as input is a very similar problem. Here, word embeddings \cite{levy2014neural} are used to encode words into vectors that can be used as input data. Following this method, we also embed our lines, but first we need to define what such a line actually consists of:
\begin{itemize}
	\item Pen State (indicator if line should be drawn or not)
	\item Position (position to move to; relative to last position)
\end{itemize}
This representation is very similar to Turtle Graphics, where a virtual pen can be moved with relative position commands. In the following, we refer to this line definition as a ``move". Additionally, we introduce a special move to indicate the end of the image, which is also added to the embedding. The stroke ending does not need an own indicator move, as it is encoded in the Pen State change.\\
\ \\

\section{Data Set Generation}
\begin{figure}[h]
	\centering
	\includegraphics[width=8cm]{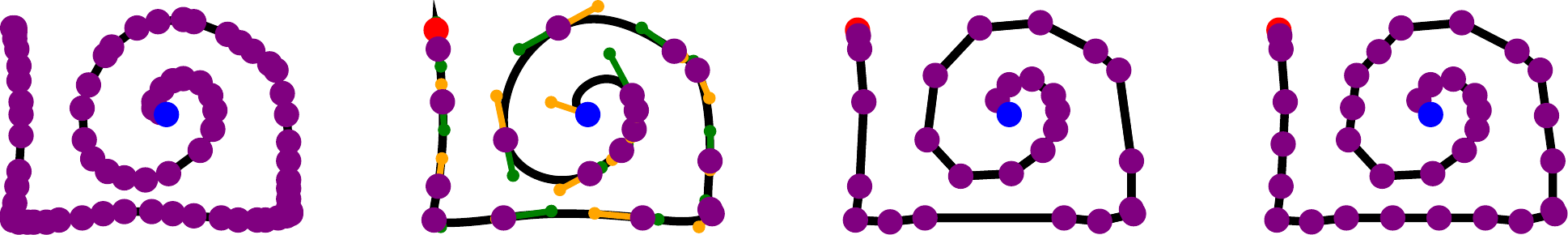}
	\caption{Simplification of a path: the first image depicts a hand-drawn stroke, where on each mouse event a new point is recorded. The next image shows a simplified version (a path), where multiple points have been substituted with a curve. In the next step, the path is converted to moves where too long moves are divided into multiple shorter moves.}
	\label{simplify}
\end{figure}

In our experiments, the initial stroke drawings are hand-drawn with a digitizer pen, from which only points are recorded (see figure \ref{simplify}). The point sequence is then simplified to remove unnecessary points and instead describe the hand-drawn stroke by few curves instead of many points. So, a sequence of as few curves as possible is fitted through the recorded pen position points with an allowed maximum error \cite{schneider1990algorithm}. The resulting sequence of curves will in the following referred to as a path. The recorded drawing is stored in this path state, because it is a better approximation as the later constructed straight moves, especially if the path is altered to create a large data set as described below.
\ \\
When learning generative models from few natural images \cite{shaham2019singan}, the images are altered into a variety of so called patches. These patches are cut out parts of the original images, which are also often slightly deformed, scaled or changed in other manners to produce a larger amount of training data as the initial images would have provided.\\
To learn a generative model on one single sketch image, we propose a similar method: the initial strokes are altered in different ways to produce a large and diverse training data set. As stroke-based images differ a lot from natural, pixel-based images, the altering methods need to be adapted accordingly. All proposed altering methods are visualized in figure \ref{manipulation} and will be described below:
\begin{figure}[h]
	\centering
	\includegraphics[width=6cm]{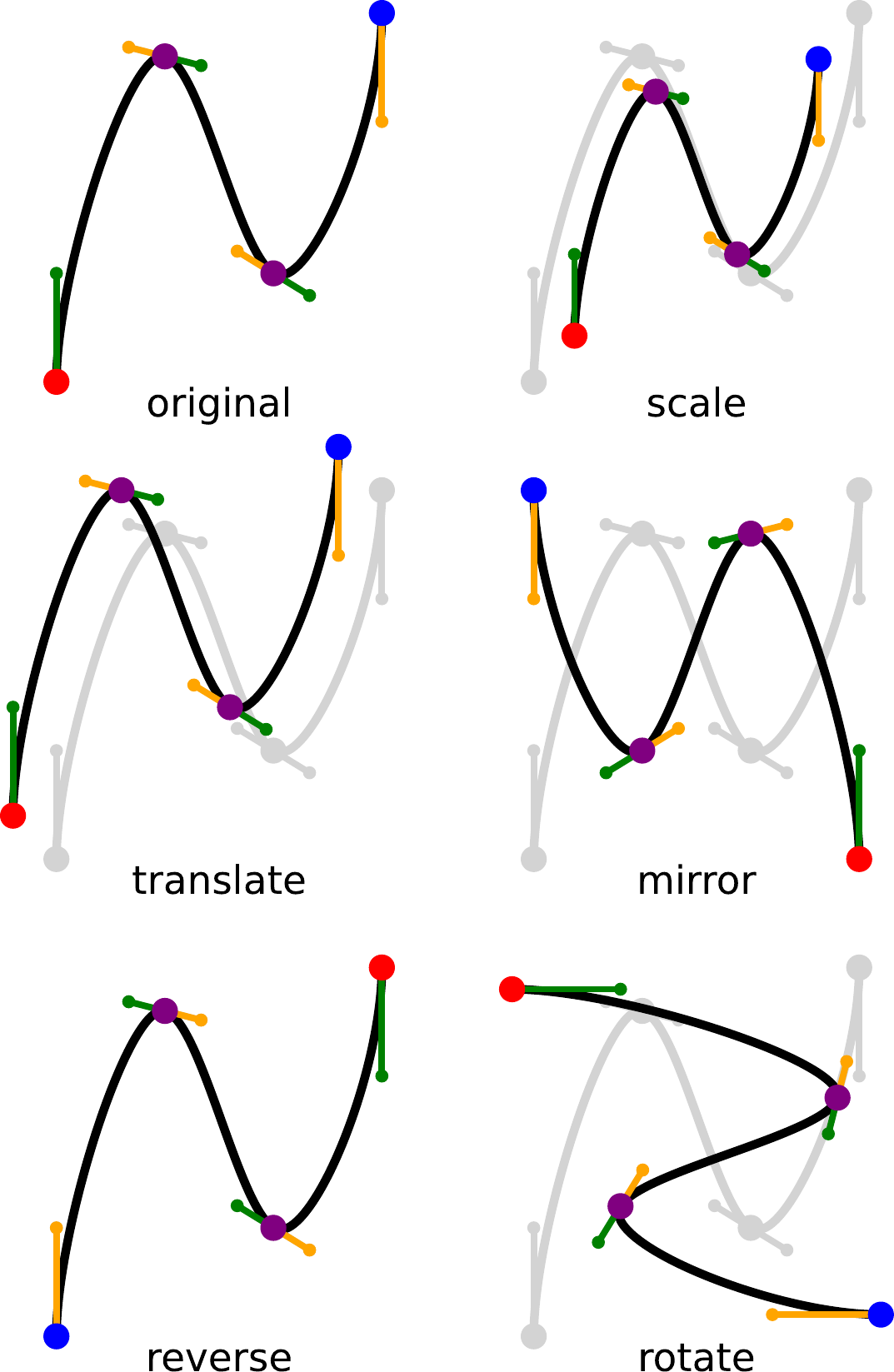}
	\caption{Five types of path manipulations used to create a large data set from one example. Translation, Rotation, Scaling and Mirroring are used on the whole stroke-based image, whereas Path Reversal is used on individual paths in the image.}
	\label{manipulation}
\end{figure}
\begin{itemize}
	\item \textbf{Translation:} The whole path-image is moved to a new position in a way, that it is still contained in the initial image boundaries.
	\item \textbf{Rotation:} The whole path-image is rotated by a random angle.
	\item \textbf{Mirror:} the path image is mirrored along an axis.
	\item \textbf{Path Reversal:} As each path consists of a list of curves with start and end points, a path has an implicit direction. In our setting, the path direction is not important, so a path can be reversed to generate new patches. The path direction might be important for other settings like sketch or handwriting classification \cite{xu2019multi}, where the stroke direction and path order are very similar in one letter. As Path Reversal is a binary state (either the path is reversed or not), this manipulation is applied with a probability of 0.5.
	\item \textbf{Scaling:} The whole path-based image is scaled to a smaller size.
\end{itemize}
After all paths have been manipulated, they are rearranged in a new order. We sort them in a greedy way by distance, so that the pen travel is as short as possible. We start at a randomly chosen path in the image. With this new path order, we add more variety in the data set while assuring that the model will learn to draw close by the last stroke. If paths would be not sorted but shuffled randomly, the resulting images will look more scattered, as new paths would appear in greater distances.\\

\section{Training}
\begin{figure*}[t]
	\centering
	\begin{minipage}[b]{0.15\textwidth}
		\centering
		\includegraphics[width=\textwidth]{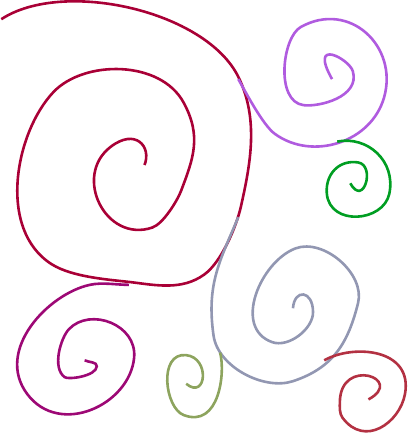}
		\caption[Network2]%
		{{\small ``curles"}}    
		\label{fig:mean and std of net14}
	\end{minipage}
	\qquad
	\begin{minipage}[b]{0.15\textwidth}  
		\centering 
		\includegraphics[width=\textwidth]{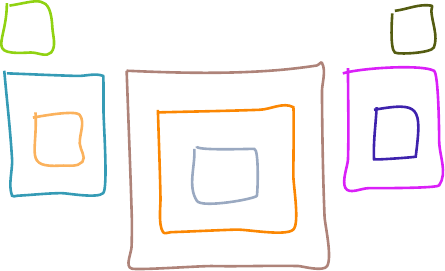}
		\caption[]%
		{{\small ``boxes"}}    
		\label{fig:mean and std of net24}
	\end{minipage}
	\qquad
	\begin{minipage}[b]{0.15\textwidth}   
		\centering 
		\includegraphics[width=\textwidth]{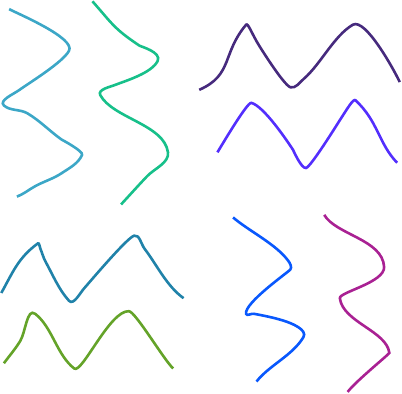}
		\caption[]%
		{{\small ``spikes"}}    
		\label{fig:mean and std of net34}
	\end{minipage}
	\qquad
	\begin{minipage}[b]{0.15\textwidth}   
		\centering 
		\includegraphics[width=\textwidth]{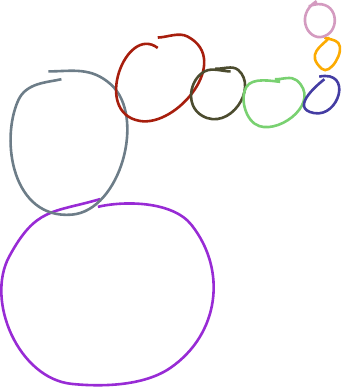}
		\caption[]%
		{{\small ``cirlces"}}    
		\label{fig:mean and std of net44}
	\end{minipage}
	\caption[ The average and standard deviation of critical parameters ]
	{\small Four recorded stroke-based images to be used in the experiments. Each stroke is colored randomly for better distinction.} 
	\label{origins}
\end{figure*}
\begin{figure}[h]
	\includegraphics[width=8cm]{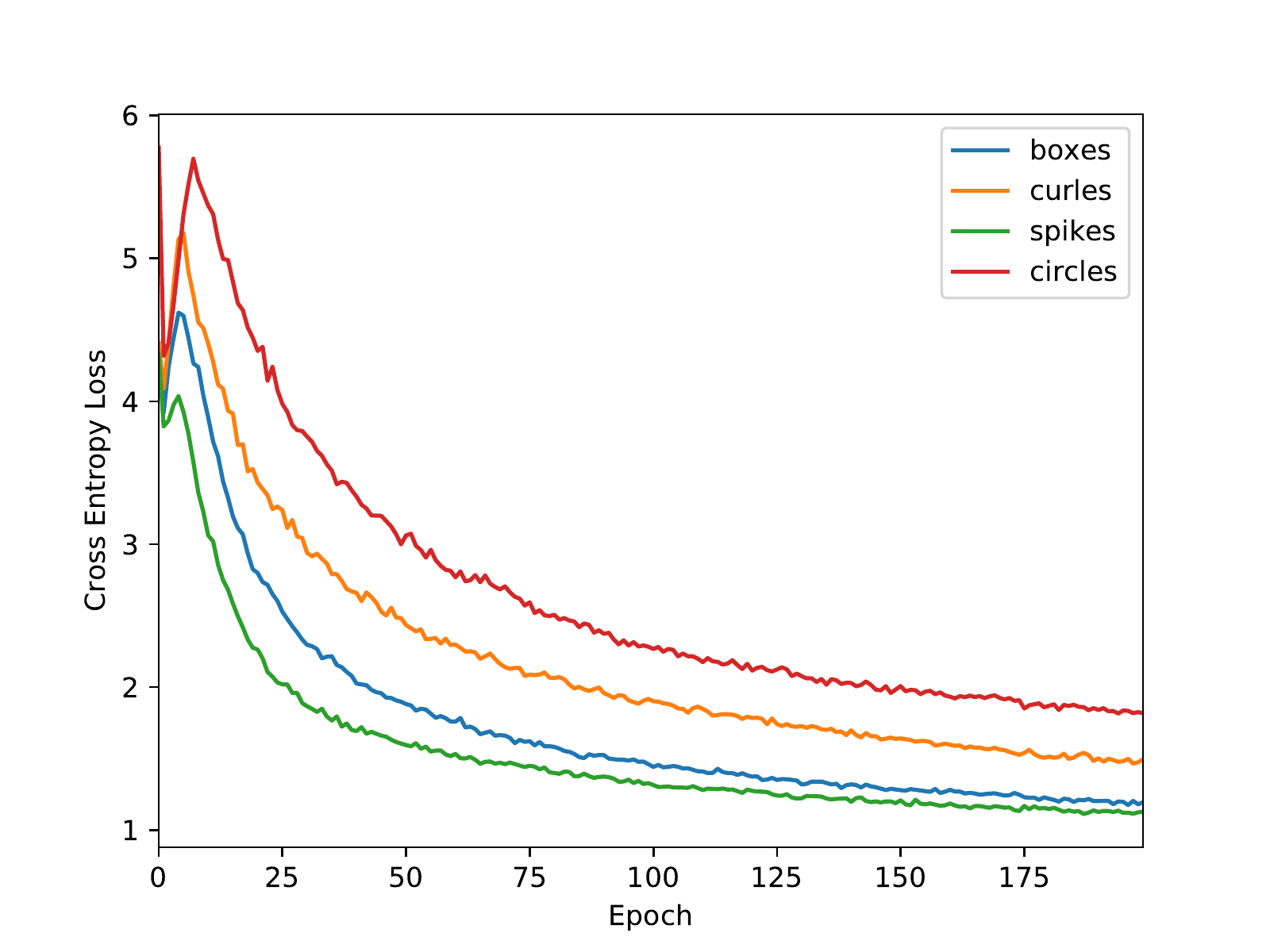}
	\caption{Loss of one unseen patch set of size 500. Loss is calculated with every model after each epoch. Models were trained with newly generated patch sets for every epoch. The falling loss curve indicates that the model learns to represent the whole patch distribution. (Absolute Loss differs because of different sequence lengths between models).}
	\label{generalization}
\end{figure}
\begin{figure}[h]
	\includegraphics[width=8cm]{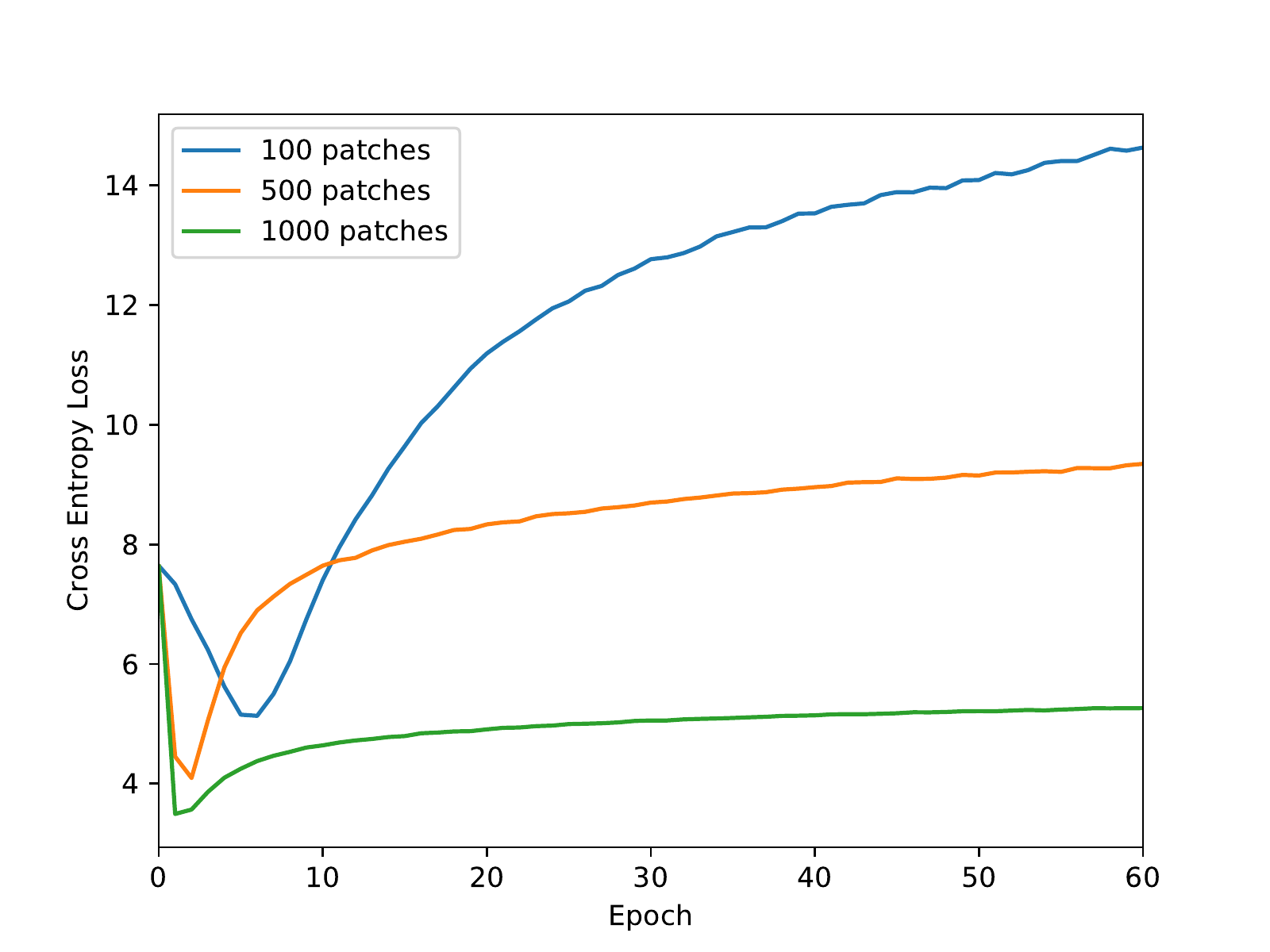}
	\caption{Loss of one unseen patch set of size 500. Loss is calculated with three versions of the ``boxes" model: one which was trained with a patch set size of 100, 500 and 1000. The rising loss curve indicates that the model overfits to the small patch sample.}
	\label{overfitting}
\end{figure}
\begin{figure}[h]
	\centering
	\includegraphics[width=7cm]{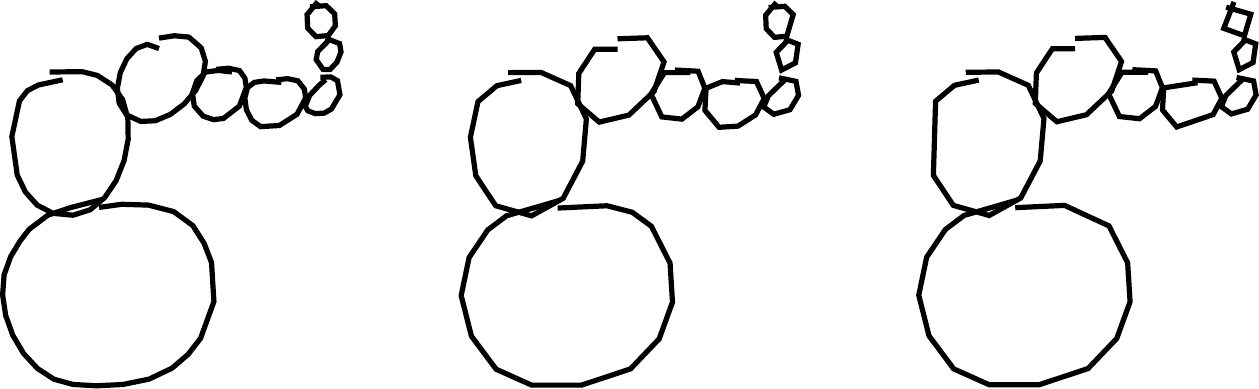}
	\caption{Curve flattening results with maximum allowed error of 1(left) to 3(right). With an higher allowed error, smaller details like narrow curves suffer and are approximated to sharp edges. }
	\label{flatten}
\end{figure}

We recorded 4 different drawings with an image boundary of 180x180 units to serve as initial stroke-based images, which can be seen in figure \ref{origins}.\\
Each model was trained for 200 epochs, where for each epoch a new data set of 500 patches is generated from the original image. These patches were then converted to moves with a maximum line length of 15 units.\\
It is important to generate a new patch data set each epoch to prevent overfitting to a small patch set. Instead, with the changing patch training data, the model sees a larger portion of the possible patchs.\\
To verify that the model learns to represent the whole patch distribution, we calculated the Cross Entropy Loss of an unseen data set of 500 patches after each training episode. The results can be seen in figure \ref{generalization}. After an initial increase, the loss sinks for each new training epoch. Because the model sees a large portion of the data set distribution, it better learns to represent the whole data set distribution.\\
If the model would only be trained on one patch data set, the model quickly begins to overfit as can be seen in figure \ref{overfitting}. Here the ``boxes" model was trained with one single patch of size 100, 500 and 1000. The plot depicts the Cross Entropy Loss of these models between an unseen patch set of size 500. The larger the training patch set is, the lower the loss. Though, the loss increases for each episode as one small patch set badly represents the whole data set distribution and the model overfits to the small sample.\\
\\
When generating moves, the approximation error for the curve flattening algorithm should not be set too high, as the resulting image quality suffers. The difference of an error between 1 and 3 can be seen in figure \ref{flatten}. With an large allowed error, especially details in narrow curves are lost. For our research we chose a maximum error value of 1.
\\
\ \\
In the training phase, we do not feed patch by patch to the Transformer. Instead we feed in a continuous stream of patches (though the input vectors are shuffled). So one input vector can contain for example the end half of a patch and the beginning halt of another patch. This way the Transformer learns to generate a stream of images. This will be helpful in the generation phase.\\
For our training, we use the Adam optimizer \cite{kingma2014adam} with the described changes in \cite{vaswani2017attention}, where the learning rate is first linearly increased for the first warm-up steps and thereafter decreased again.\\
As a loss function, we use Cross Entropy Loss.\\
For all of our experiments, we used the following Transformer settings:
\begin{itemize}
	\item Batch Size: 200
	\item Sequence Length: max. move length of recorded image
	\item Hidden Embedding Size: 52
	\item Decoder Layers: 6
	\item Attention Heads: 4
	\item Feed Forward Size: 2048
\end{itemize}

\section{Inference and Sampling}
\begin{figure}[p]
	\centering
	\includegraphics[width=8cm]{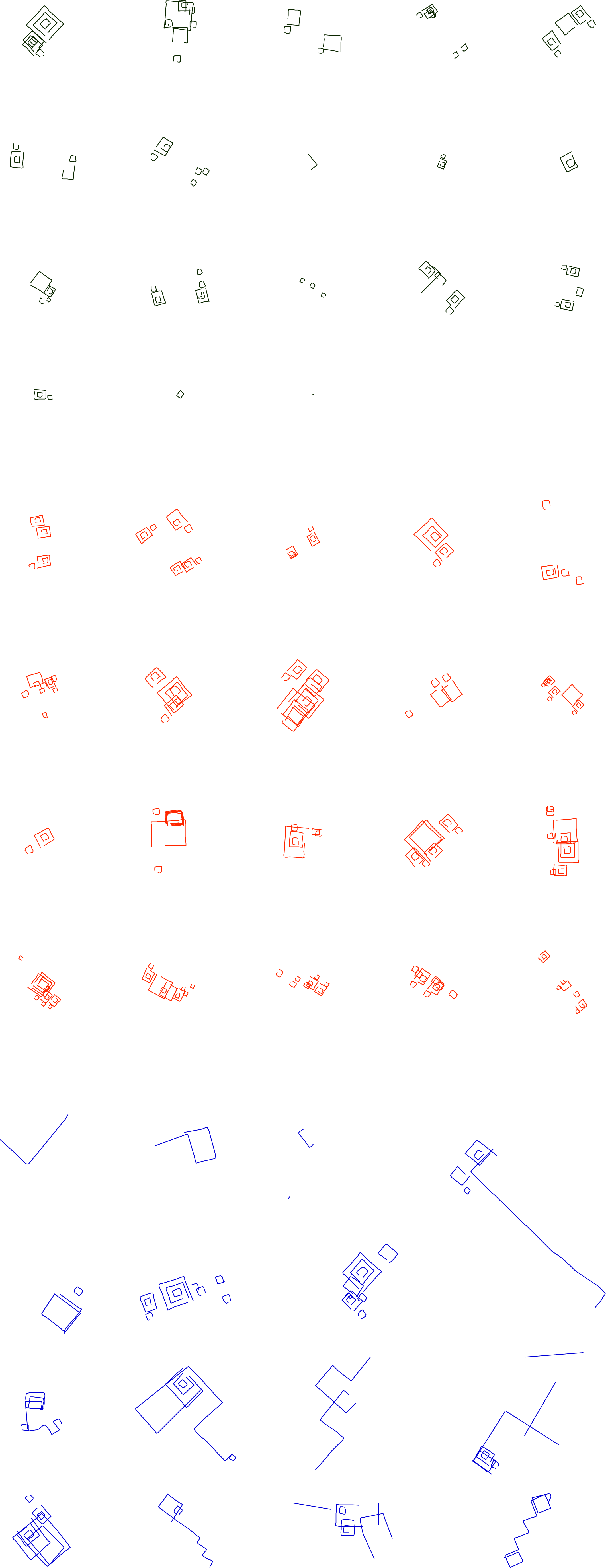}
	\caption{Group of sampled sketches, differing in the inizialization vector length: 1 (black), half of sequence length (red) and full sequence length (blue). }
	\label{init}
\end{figure}
\begin{figure*}[]
	\centering
	\begin{minipage}[b]{0.35\textwidth}
		\centering
		\includegraphics[width=\textwidth]{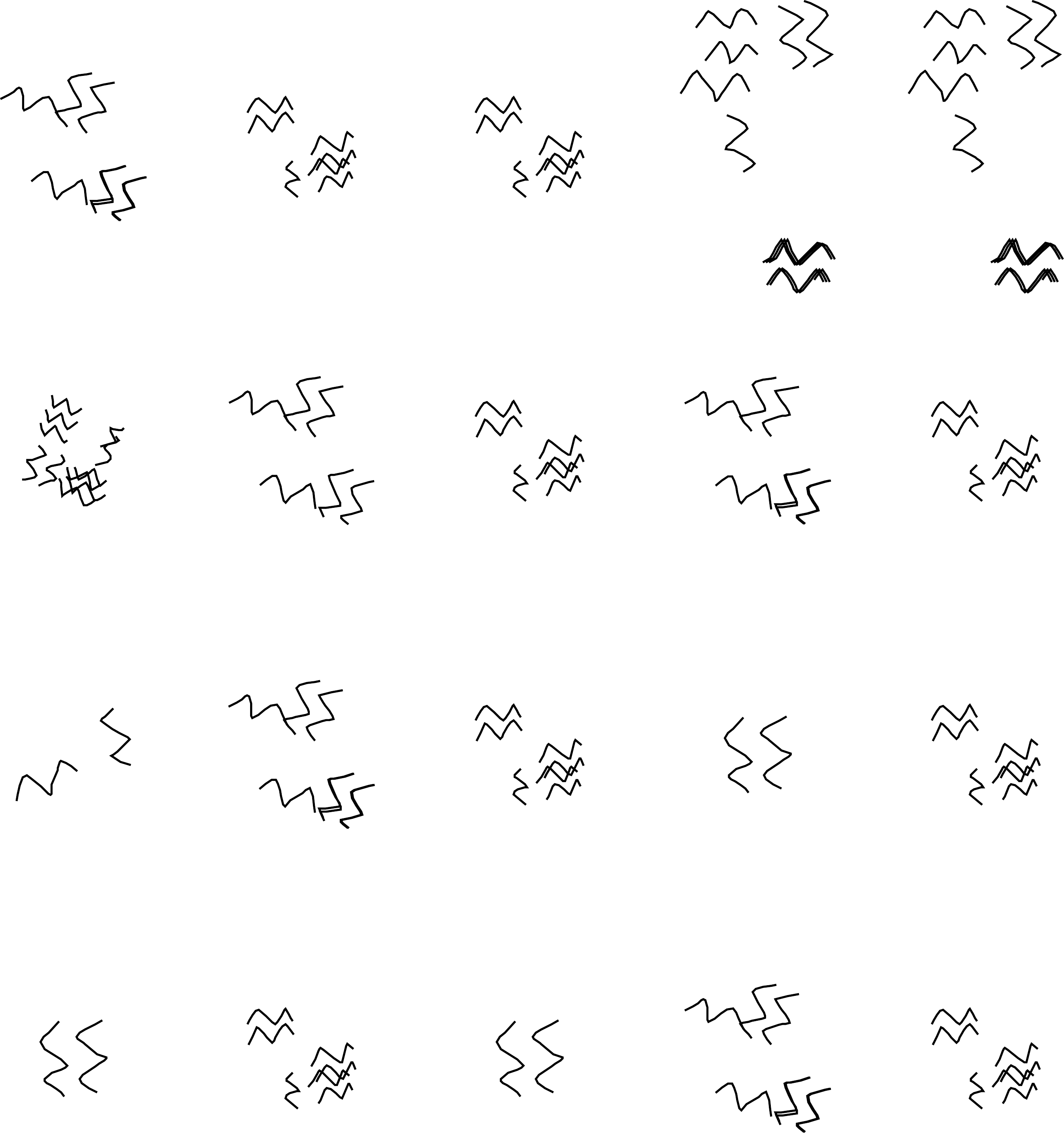}
	\end{minipage}
	\qquad
	\begin{minipage}[b]{0.35\textwidth}  
		\centering 
		\includegraphics[width=\textwidth]{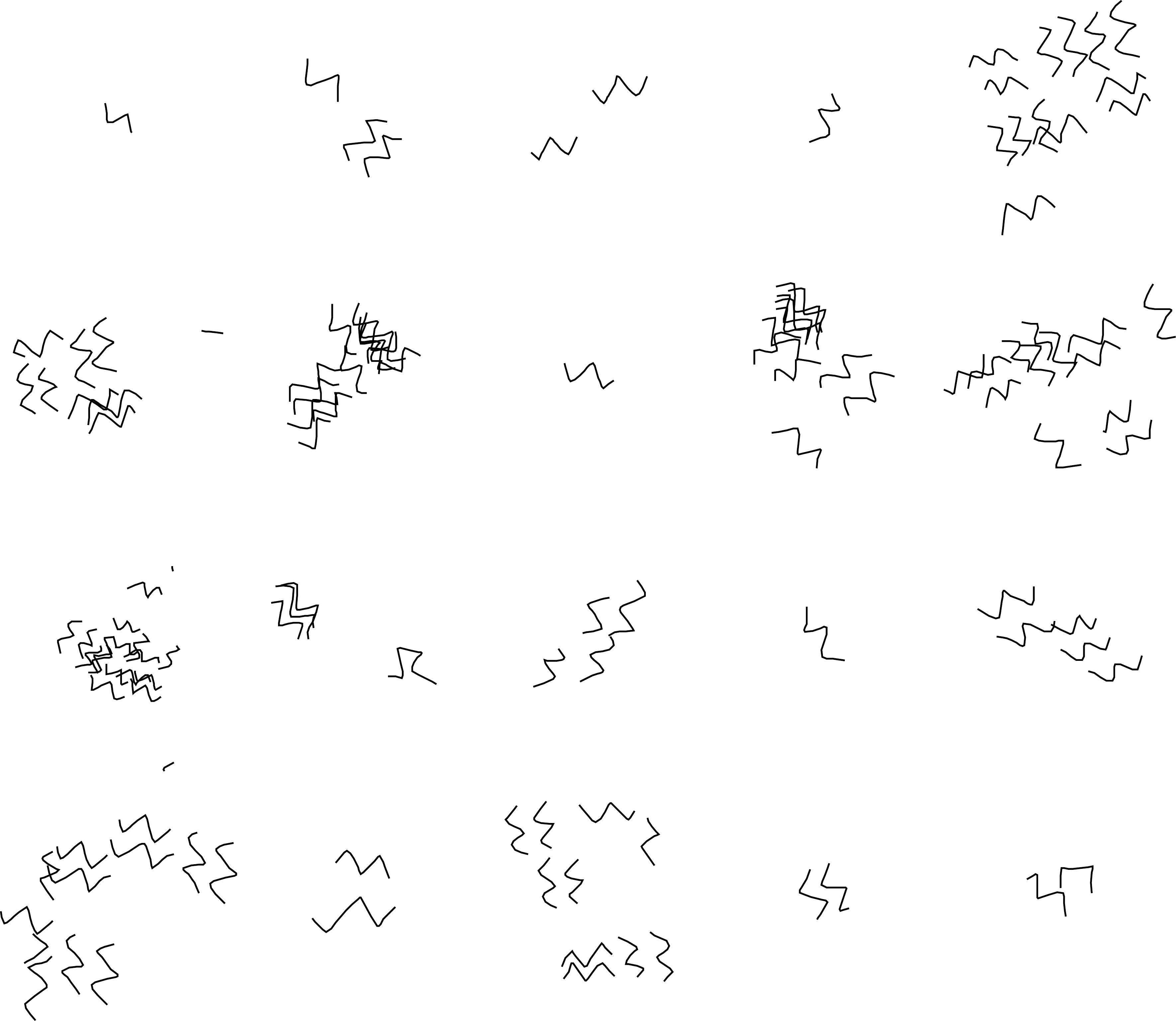}
	\end{minipage}
	\\
	\vspace{0.5cm}
	\begin{minipage}[b]{0.35\textwidth}   
		\centering 
		\includegraphics[width=\textwidth]{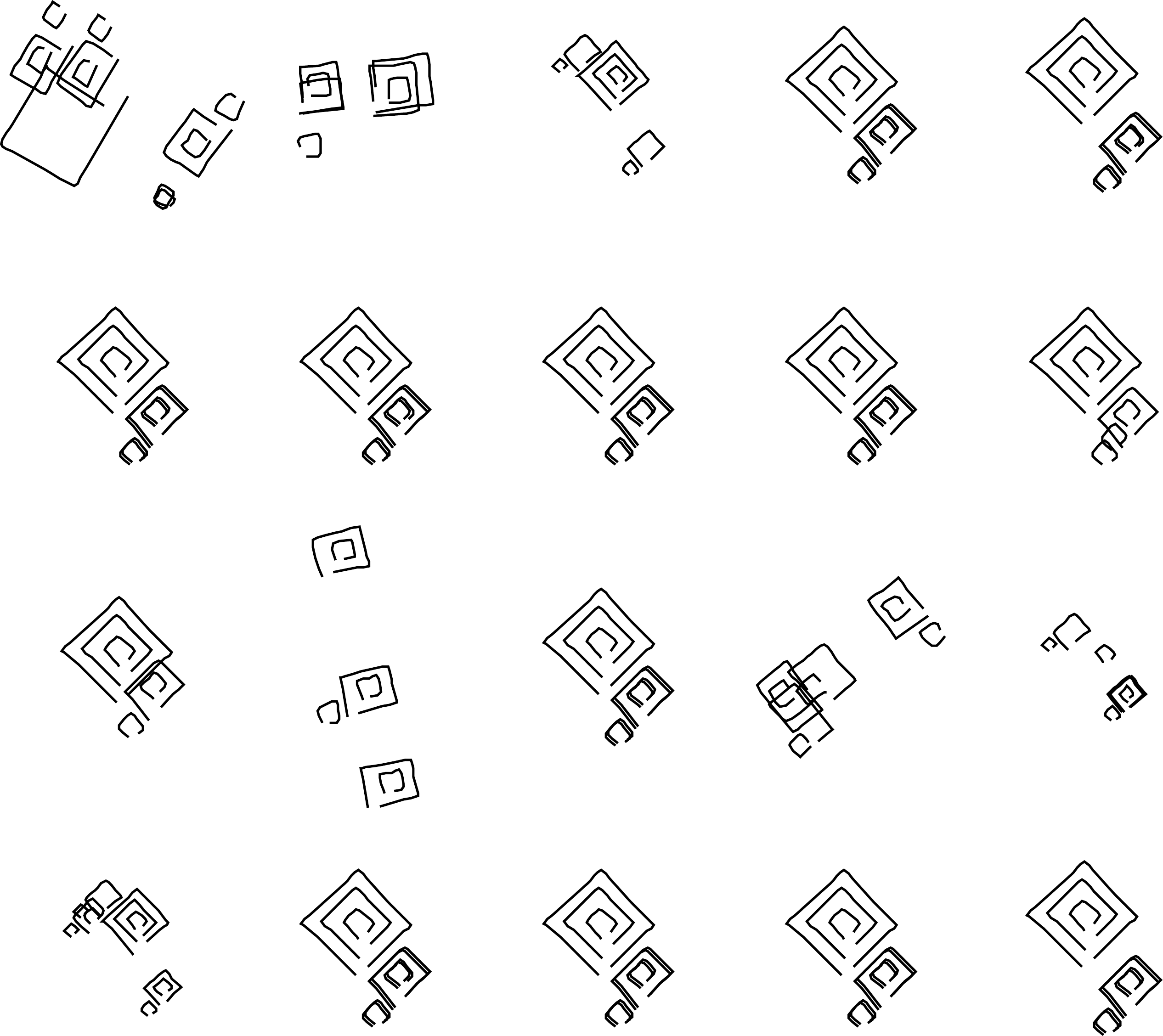}
	\end{minipage}
	\qquad
	\begin{minipage}[b]{0.35\textwidth}   
		\centering 
		\includegraphics[width=\textwidth]{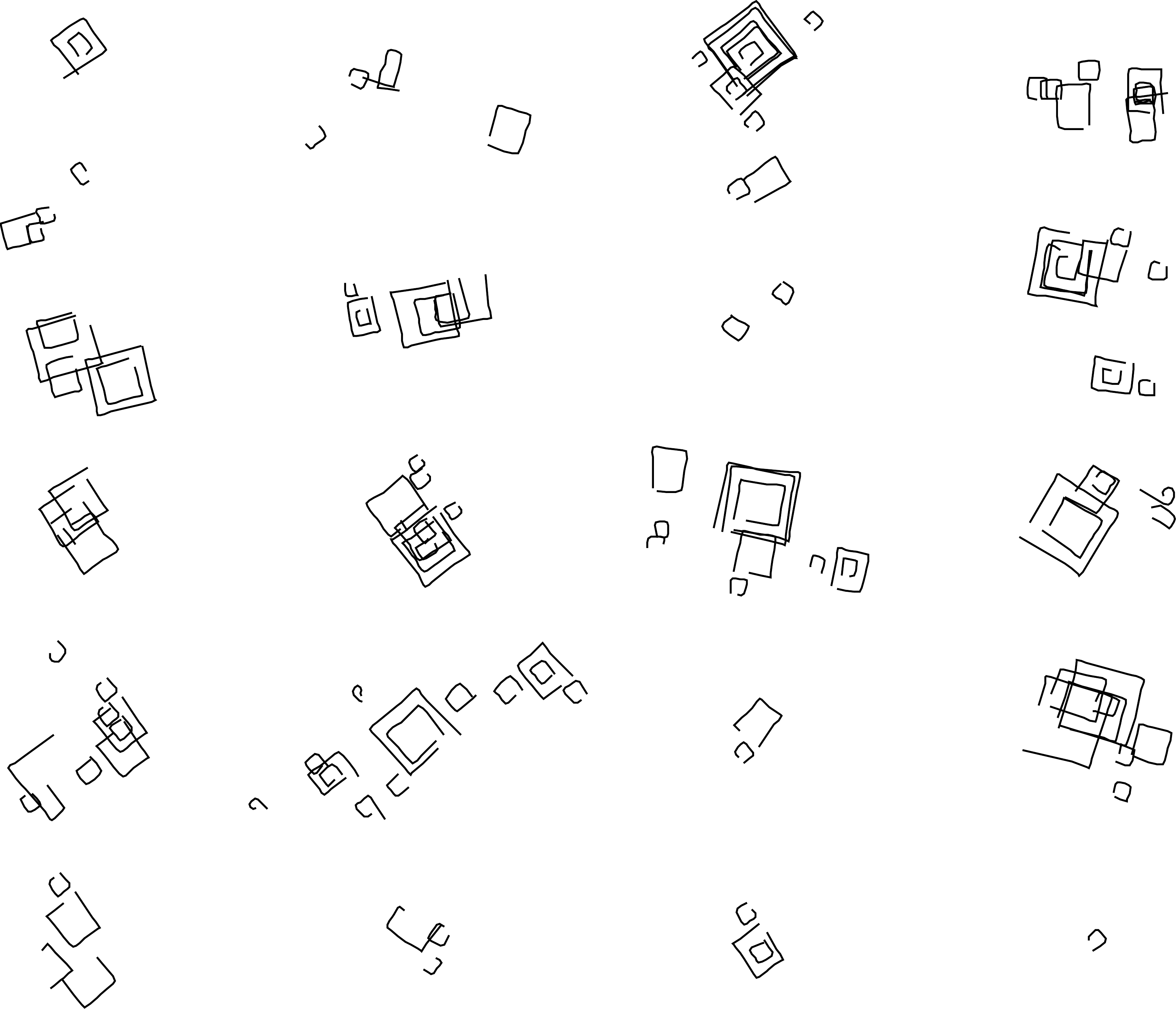}
	\end{minipage}
	\\
	\vspace{0.5cm}
	\begin{minipage}[b]{0.35\textwidth}
		\centering
		\includegraphics[width=\textwidth]{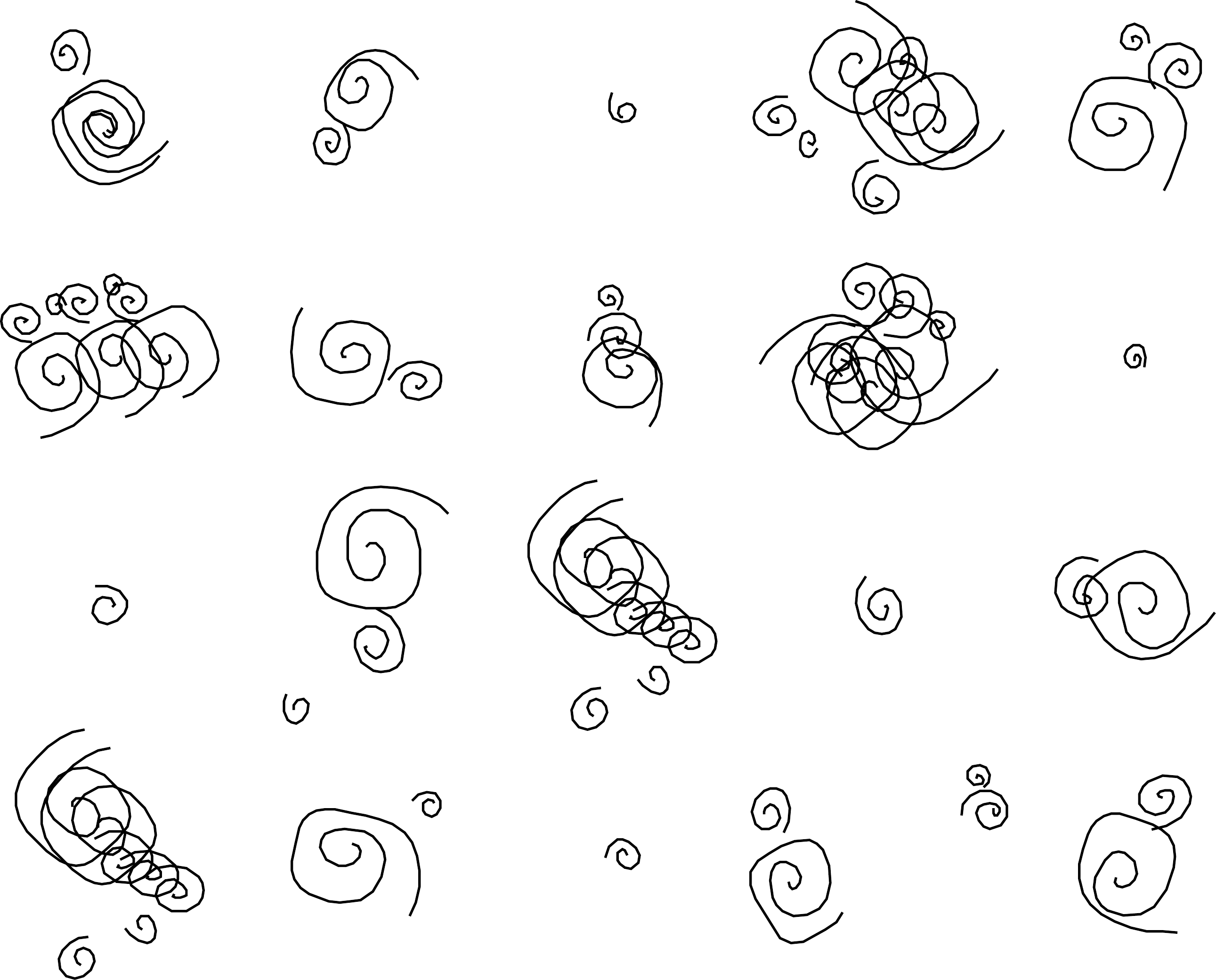}
	\end{minipage}
	\qquad
	\begin{minipage}[b]{0.35\textwidth}  
		\centering 
		\includegraphics[width=\textwidth]{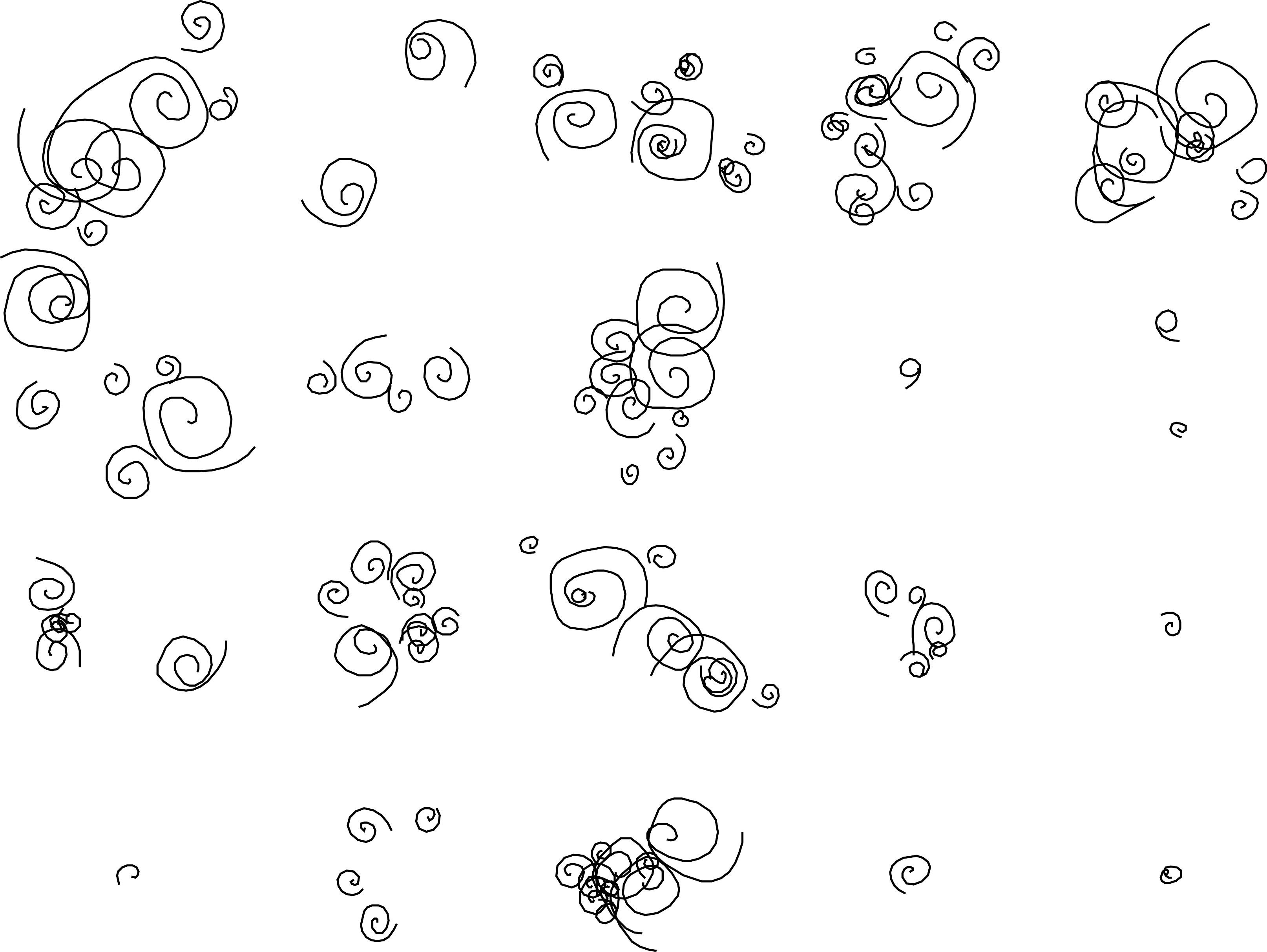}
	\end{minipage}
	\\
	\vspace{0.5cm}
	\begin{minipage}[b]{0.35\textwidth}   
		\centering 
		\includegraphics[width=\textwidth]{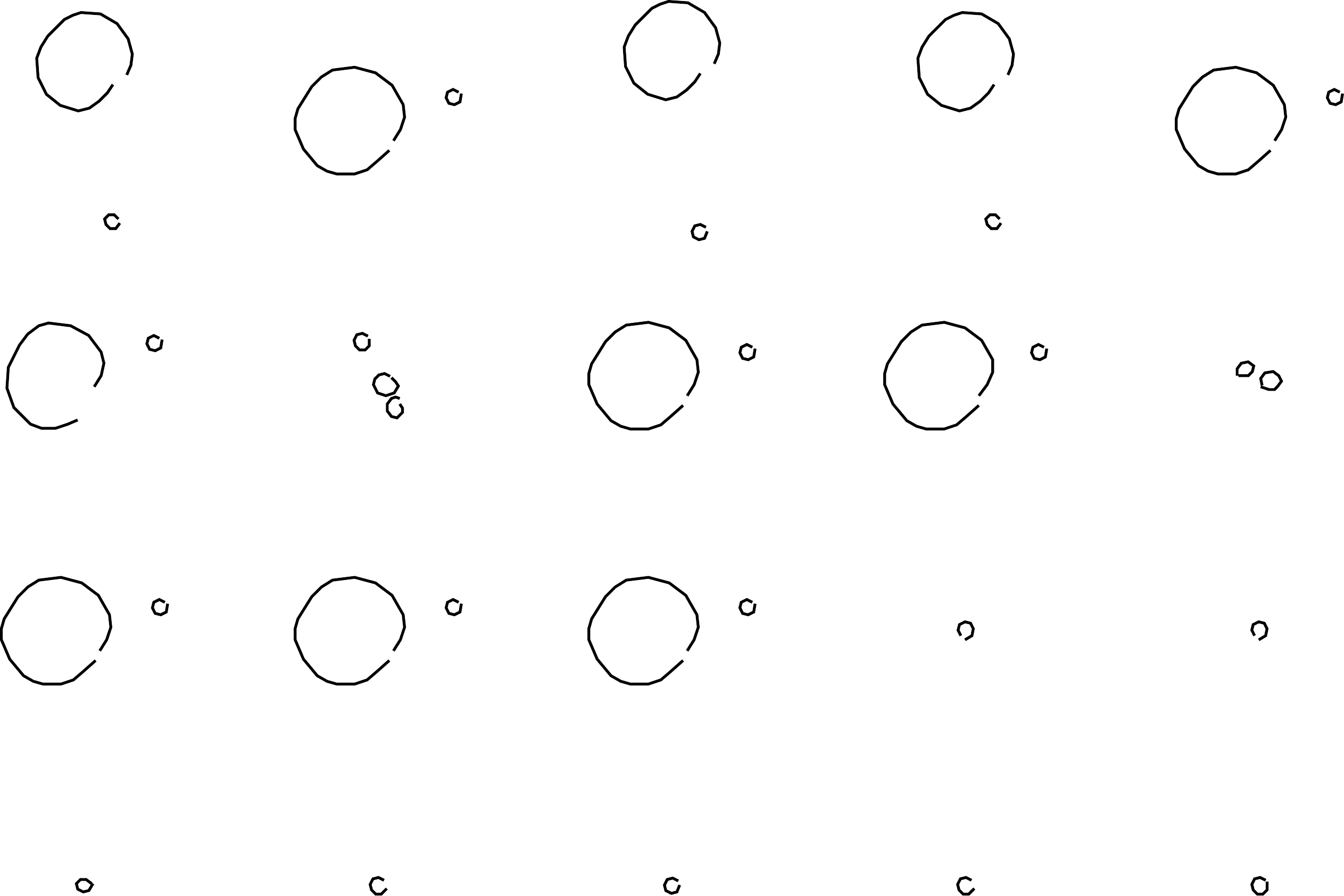}
	\end{minipage}
	\qquad
	\begin{minipage}[b]{0.35\textwidth}   
		\centering 
		\includegraphics[width=\textwidth]{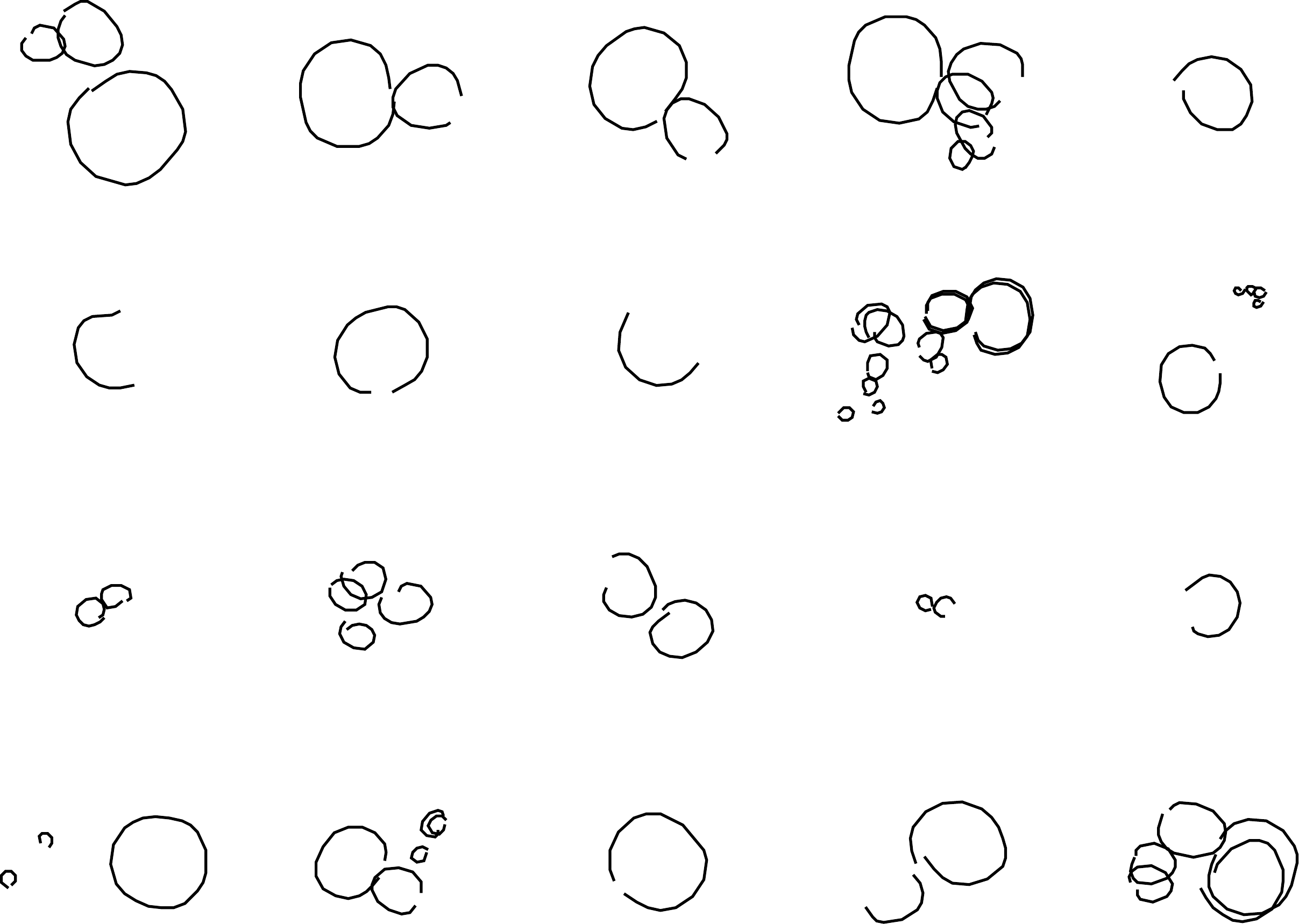}
	\end{minipage}
	\caption{Groups of top-k sampled sketches from all four models. Left: k=0, right: k=10.}
	\label{topk}
\end{figure*}
The Transformer can only predict one new move dependent on the previous moves. So the Transformer needs at least one move to start generating. We experimented with different lengths of random input vectors as initialization. These random vectors always ended with the ``image-end" move to trigger the generation of a new image. The random initialization vector is of course discarded and not shown in the drawings. Figure \ref{init} shows the results of three different initialization vector lengths. The black group was generated only by passing the ``image-end" move to the Transformer. The images are often rather short (in sense of stroke count) and rarely show strokes that don't seem to fit to the original image. Also some empty images occurred.\\
The red block was generated with a random initialization vector of half the sequence length. The images show a high variety and seem to fit well to the original image.\\
The blue group on the bottom of figure \ref{init} was generated with an initialization vector of the full sequence length. Here, the model seems ``confused" by the random input and produces a large variety of long or unfitting strokes.\\
We suggest to use an initialization vector in the range of half the sequence length to create enough randomness for interesting results but not to confuse the model too much.\\
\ \\
As a sampling strategy we use top-k sampling. Here, only the best $k$ predictions are considered and are then chosen according to their probability. If $k=1$ top-k sampling is equivalent to a greedy approach where every time only the best prediction is chosen. Figure \ref{topk} shows results from all trained models with two different $k$ values: the left side is a greedy sampling with $k=1$ where the right side shows a sampling with $k=10$.\\
The mayor difference in both sets is the image variety. With the greedy approach, the model often falls back to a small set of images. But these images fit well to the original image style. Though when generating, we experienced that the Transformer got stuck in generative loops, where it would draw one shape indefinitely.\\
The images sampled with top-10 show a larger variety and also fit well to the original image, except some individual strokes. Over all these generated images look a bit more chaotic and also show new shapes that are not part of the original image, but fit the style (f.e. double-curled lines in the ``curles" image).\\
In Natural Language Processing, the sampling technique is often adjusted to the generation task. So in case of generating stroke images, it might also be best to adjust the sampling algorithm to the intended purpose.\\
\begin{figure}[h]
	\centering
	\includegraphics[width=8cm]{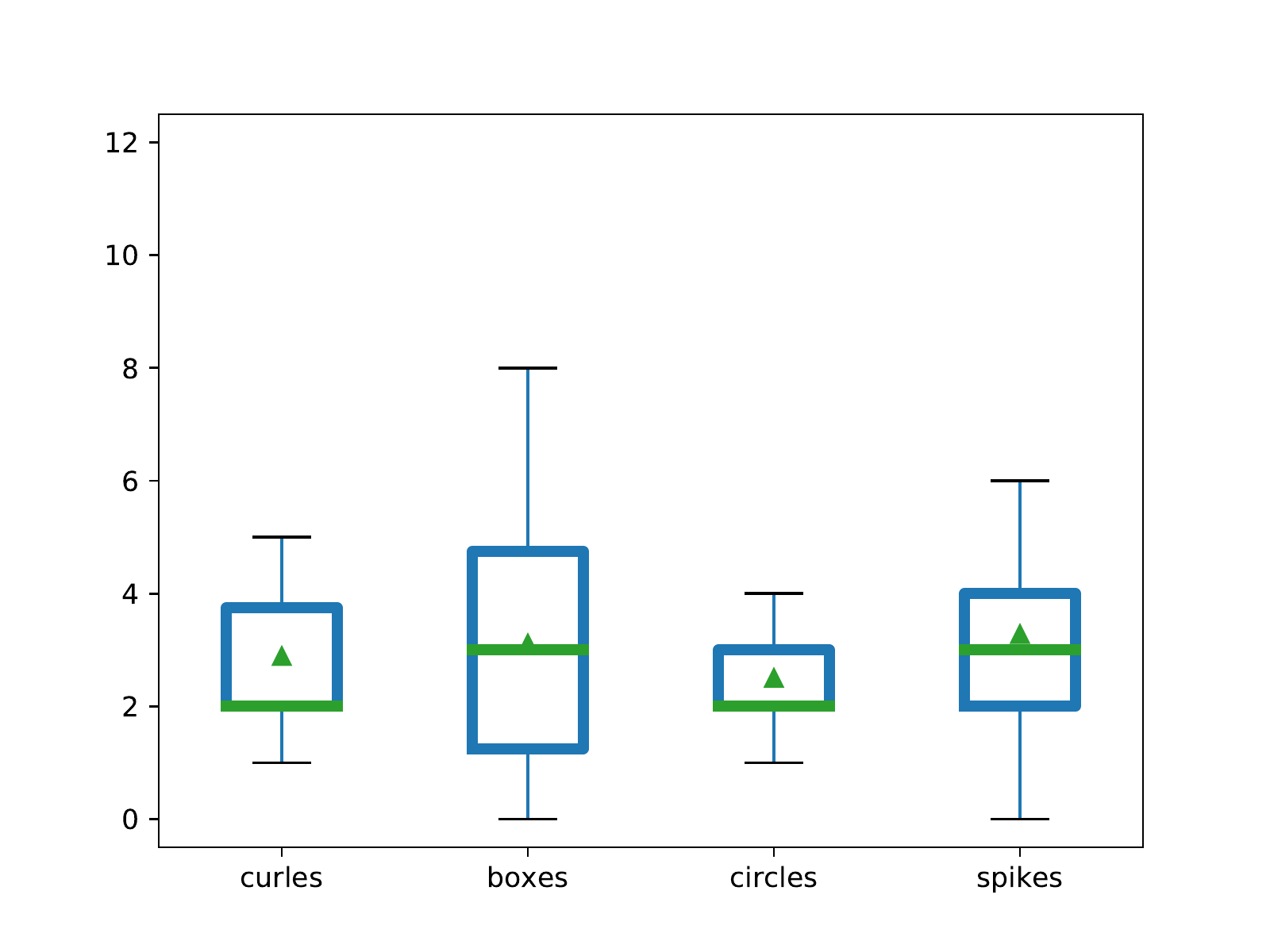}
	\caption{18 participants chose out of a set of 12 pattern describing adjectives, which fit best for the original image and a set of generated images. Plot shows the error distribution grouped by model. The average error is 2.94.}
	\label{boxplot}
\end{figure}\\
To evaluate if the generated images preserve the style of the original sketch, we asked 18 participants to choose from a set of 12 adjectives (parallel, ordered, stacked, chaotic, curly, straight, symmetric, repetitive/rhythmic, tangled, circular, jagged, nested) which ones best describe the original image and a set of generated images. The adjectives were chosen in a way that in combination they are able to describe a large variety of pattern. The question order was counterbalanced to avoid bias. We calculated the difference in chosen adjectives for each person. Figure \ref{boxplot} shows the results for each model: the mean error over all models is 2.94 from a maximum possible error of 12. From the results we can see, that some models better preserve the style than others. The ``boxes" and ``spikes" models most likely perform worse in this test, because the trained models seem to be good in replicating drawing shapes, but not in positioning them. And these two model's main features lie in the stroke positions (for example the ``boxes" input image is symmetric).\\
Depending on the model, an average of two to three attributes changed between the assessment of the original image and the generated images. So the majority of attributes is shared between both images, which allows the conclusion that the generated model does not perfectly preserve the style but learned to replicate the large majority of attributes.

\section{Conclusion}
In our work, we showed that a Transformer neural net can be used to learn and generate stroke-based images from one single input image. We generated large training data sets from one image, using different path altering methods. Four hand-drawn image samples were recorded and used to generate our training data sets. We proved that is is essential to train the Transformer with a changing subset of training data in each epoch. This training data rotation prevents the model from overfitting to a too small sample.\\
\ \\
We compared different initialization vector lengths and found that a lengths of half the Transformer sequence length gives the best results without confusing the model. We also compared different sampling parameters in top-k sampling. We could observe that a greedy sampling method like $k=1$ results in a very low variety in images which fit very well to the original image. Increasing $k$ also increses variety but also introduces strokes that do not fit well to the original image's style. Therefore it is important to choose $k$ depending on the application needs.\\
Finally we evaluated the style preservation capabilities of our models in an assessment of 18 participants. The results showed that most of the style attributes are learned by the model, though style fitting stroke placement was not always achieved.\\
\ \\
In our future research we want to expand these generative methods to create larger path-based pattern images from one input image. For this it might be interesting to use hierarchical approaches, as they have been successfully used in other domains like dialogue generation \cite{serban2017hierarchical}. An hierarchical approach might be helpful to give the neural net an overview of already generated paths and their places beyond the sequence length memory.\\
Another interesting field is Co-Creative Design \cite{guzdial2018co}, where a user is cooperating with an artificial agent to support him in a design task. Here, a stroke-based image representation can be very useful, especially in the domain of digital fabrication or Casual Creators. It will be interesting to explore these applications and domains in future research.

\bibliographystyle{iccc}
\bibliography{icccbib}

\end{document}